\documentclass[conference]{IEEEtran}
\usepackage[space]{cite}
\usepackage{amsmath,amssymb,amsfonts}
\usepackage{graphicx}
\usepackage{multirow}
\usepackage{amsthm}
\usepackage{algorithm}
\usepackage{algpseudocode}
\usepackage{url}
\usepackage{textcomp}
\usepackage{xcolor}
\usepackage{subcaption}
\usepackage{fancyhdr}
\fancypagestyle{body}{
    \fancyhead{}
    
    \fancyfoot{}
}
\def\BibTeX{{\rm B\kern-.05em{\sc i\kern-.025em b}\kern-.08em
    T\kern-.1667em\lower.7ex\hbox{E}\kern-.125emX}}

\DeclareMathOperator*{\argmin}{arg\,min}

\begin{document}
\title{Continual Learning with Gated Incremental Memories for sequential data processing} 

\author{\IEEEauthorblockN{Andrea Cossu}
\IEEEauthorblockA{\textit{Computer Science Dept.} \\
\textit{University of Pisa}\\
Pisa, Italy \\
andrea.cossu@sns.it}
\and
\IEEEauthorblockN{Antonio Carta}
\IEEEauthorblockA{\textit{Computer Science Dept.} \\
\textit{University of Pisa}\\
Pisa, Italy \\
antonio.carta@di.unipi.it}
\and
\IEEEauthorblockN{Davide Bacciu}
\IEEEauthorblockA{\textit{Computer Science Dept.} \\
\textit{University of Pisa}\\
Pisa, Italy \\
bacciu@di.unipi.it}
}
\maketitle
\thispagestyle{fancy}
\begin{abstract}
The ability to learn in dynamic, nonstationary environments without forgetting previous knowledge, also known as Continual Learning (CL), is a key enabler for scalable and trustworthy deployments of adaptive solutions. While the importance of continual learning is largely acknowledged in machine vision and reinforcement learning problems, this is mostly under-documented for sequence processing tasks. This work proposes a Recurrent Neural Network (RNN) model for CL that is able to deal with concept drift in input distribution without forgetting previously acquired knowledge. We also implement and test a popular CL approach, Elastic Weight Consolidation (EWC), on top of two different types of RNNs. Finally, we compare the performances of our enhanced architecture against EWC and RNNs on a set of standard CL benchmarks, adapted to the sequential data processing scenario. Results show the superior performance of our architecture and highlight the need for special solutions designed to address CL in RNNs.

\end{abstract}


\section{Introduction}\label{sec:intro}
Dynamic environments are often subjected to the concept drift phenomenon \cite{widmer_learning_1996, gama_survey_2014, tsymbal_problem_2004, lu_learning_2019, ditzler_learning_2015} which reflects substantial changes in the data generating process and the corresponding predictions. More formally, given an unknown time-dependent joint probability $p_t(y, \mathbf{x})$ over data $\mathbf{x}$ and target $y$, concept drift can affect both the evidence $p_t(\mathbf{x})$ and the conditional distribution $p_t(y | \mathbf{x})$. Following the definitions presented in \cite{ditzler_learning_2015}, this paper addresses the problem of learning in dynamic environments in which the evidence $p_t(\mathbf{x})$ exhibit instantaneous drifts without any time limit imposed on it, leading to a \textit{permanent, abrupt concept drift}. \\
We will refer to a specific objective (e.g. learn to classify MNIST digits) as \textit{task}. An \textit{input distribution}, instead, will generate data related to a particular task (e.g. subsets of MNIST digits). Each task is associated to multiple input distributions (also called subtasks) which altogether define a dynamic environment in which the model is trained. We will deal with input sequences in the form  $\mathbf{x} = [x_i]_{i=1,...,T}$, $x_i \in \mathbb{R}^d$ and scalar targets $y \in \mathbb{R}$. \\

In the presence of dynamic environments with recurring concepts drifts \cite{widmer_learning_1996} it is useful to design models that are able to recall and exploit previously acquired information. Unfortunately, continuous plasticity of internal representations under drifting task distributions is widely known to suffer from negative interference between the tasks that are incrementally presented to the model, yielding to the well known stability-plasticity dilemma \cite{parisi_continual_2019,zhou_online_2012} of connectionist models. The result is that the models catastrophically forget previously acquired knowledge \cite{robins_catastrophic_1995, french_catastrophic_1999} as new tasks become available.\\ 
The problem of catastrophic forgetting is the main focus of Continual Learning (CL), defined as \textit{``the unending process of learning new things on top of what has already been learned"} \cite{ring_recurrent_2011}. \\
In the CL scenario, a learning model is required to incrementally build and dynamically update internal representations as the distribution of tasks dynamically changes across its lifetime. Ideally, part of such internal representations will be general and invariant enough to be reusable across similar tasks, while another part should preserve and encode task-specific representations. \\
While current trends in CL put particular emphasis on computer vision applications or reinforcement learning scenarios, sequential data processing is rarely taken into consideration (see Section \ref{sec:related}). However, sequential data are heavily used in several fields like Natural Language Processing, signal processing, bioinformatics, and many others. In this context, Recurrent Neural Networks (RNNs) have the ability to develop neural representations that capture the history of their inputs. Learning proper memory representations is a major challenge for RNNs. In addition, in a CL setting, RNNs have to deal with drifts in task distributions which can greatly affect their capability of developing robust and effective memory representations. \\

We provide a threefold contribution to the discussion concerning CL in sequential data processing. First, we define a new dynamic approach, named Gated Incremental Memory (GIM), that imbues RNN architectures with CL skills by incrementally adding new modules to capture the drifts in input distribution while avoiding catastrophic forgetting. GIM leverages autoencoders to automatically recognize input distributions and to select the correct module to process the sequence. Second, we apply Elastic Weight Consolidation (EWC) \cite{kirkpatrick_overcoming_2017}, a popular CL approach, on top of two different RNNs. At the best of our knowledge, we are the first to experiment with EWC on RNNs. \\
Third, we test the performances of GIM against EWC and standard RNNs on three benchmarks originally introduced here by adapting traditional CL tasks to the sequential case.
The results of our empirical analysis confirm the advantages of using our enhanced architecture over standard recurrent models. Such advantages are particularly clear when testing enhanced architecture on old distributions, since it successfully prevents forgetting. These results highlight some of the key differences between feedforward and recurrent CL techniques and pinpoint the need for solutions specifically designed for recurrent architectures. 

\section{Related Work} \label{sec:related}
\pagestyle{body}
\cfoot{\thepage}
Learning in dynamic environments in the presence of concept drift has received much attention in the literature. Concept drift could be generated by hidden contexts \cite{harries_extracting_1998}, whose detection would result in drastic performance improvements. Predictions on incoming input patterns can be performed relying on a window of recently encountered instances (\textit{instance selection}), thus accounting for drifts in the underlying distribution, like in FLORA systems \cite{widmer_learning_1996}. Similarly, \textit{instance weighting} systems \cite{klinkenberg_learning_2004} exploit weights on the input patterns which can inform predictions based on their similarity to specific concepts or simply their time obsolescence. Similar to the approach proposed in this paper, \textit{ensemble methods} associate an expert to each (or to a group of) concepts and combine their predictions into the final answer \cite{schlimmer_incremental_1986,stanley_learning_2001}.\\
CL explores the problem of concept drift from a slightly different perspective, by focusing on how to avoid catastrophic forgetting on old distributions, while at the same time fostering learning of incoming data. The aspect of forgetting is peculiar of CL and it is at the center of our work.\\
CL literature mostly focuses on computer vision and reinforcement learning applications, with approaches ranging from regularization methods \cite{kirkpatrick_overcoming_2017, zenke_continual_2017}, to dual models \cite{hinton_using_1987, french_pseudo-recurrent_1997}, to dynamic architectures \cite{rusu_progressive_2016, yoon_lifelong_2018}. \\
The first attempt to deal with sequential processing in CL was presented in \cite{ans_dual-network_2002}, where the authors introduced a dual model rehearsed with pseudopatterns and trained to reconstruct the next element of a sequence (i.e. sequence modeling). More recently, RNNs have been exploited in combination with other techniques, such as Fixed Expansion Layer \cite{coop_mitigation_2013}, external growing memories \cite{asghar_progressive_2018}, Reservoir Computing \cite{tetko_continual_2019} and backpropagation-free learning \cite{ororbia_continual_2019}. \\
Evaluations of the performances of standard RNNs in CL are provided in \cite{sodhani_training_2018} and \cite{schak_study_2019}. However, while the latter provides no solutions to the problem of forgetting, the former introduces an effective, but rather complex, recurrent architecture. Instead, in addition to forgetting analysis, we propose a solution that mitigates its effects, while at the same time keeping our model simple enough to favor reproducibility as well as further extensions. \\
Our dynamic RNN architecture is inspired by Progressive networks \cite{rusu_progressive_2016}, a popular CL approach used for feedforward networks that deals with drifts in the input distribution by dynamically expanding the existing model. In addition, we leverage gating autoencoders, introduced in \cite{aljundi_expert_2017} for feedforward architectures, to remove the need to know task identity at test time. \\

\section{Gated Incremental Memories for continual learning with recurrent neural networks} \label{sec:approach}

In this section, we introduce Gated Incremental Memory (GIM), a novel CL architecture designed for recurrent neural models and sequential data. In particular, we show how GIM can be obtained by combining a recurrent version of the Progressive network \cite{rusu_progressive_2016} and a set of gating autoencoders \cite{aljundi_expert_2017} to avoid, at test time, any explicit supervision about subtask labels. 
In the following, we denote an entire sequence with bold notation (e.g. $\mathbf{x}$), and a single vector with plain formatting (e.g. $x_i$). \\

\subsection{Recurrent Neural Networks}
The proposed approach is independent of the underlying recurrent architecture. To highlight the generality of our approach, we focus our study on two different classes of RNNs, using either gated and non-gated approaches. Gated models, like LSTM \cite{hochreiter_long_1997} and GRU \cite{cho_properties_2014}, leverage adaptive gates to enable selective memory cell updates. In our analysis, we consider LSTM as a representative of gated architectures, given its popularity in literature and its state-of-the-art performances in several sequential data processing benchmarks. Non-gated approaches rely on different mechanisms to solve the vanishing gradient problem, like parameterizing recurrent connections with an orthogonal matrix \cite{mhammedi_efficient_2017}. In our analysis, we consider the Linear Memory Network (LMN) \cite{bacciu_linear_2019} as a representative of non-gated approaches. LMNs leverage a conceptual separation between a nonlinear feedforward mapping computing the hidden state $h_t$ and a linear dynamic memory computing the history dependent state $h_t^m$. Briefly, in formulas: 
\begin{align*}
&h_t = \sigma(W^{xh} x_t + W^{mh} h_{t-1}^m) \\
&h_t^m = W^{hm} h_t + W^{mm} h_{t-1}^m \\
\end{align*}
where $x_t$, $h_t$ and $h_t^m$ are the input vector, the functional component activation and the memory state at time $t$. The memory state $h_t^m$ is the final output of the layer, which is given as input to subsequent layers. It is then useful to study how both models behave in CL environments and how their different memory representations respond to phenomena like drifting tasks distribution, eventually resulting in catastrophic forgetting. \\

\subsection{Elastic Weight Consolidation}
Elastic Weight Consolidation (EWC) \cite{kirkpatrick_overcoming_2017} is one of the most popular CL method. EWC mitigates forgetting by preventing large changes in those parameters which have been recognized important when learning previous distributions. Hence, in the case of recurring drifts, the model will still be able to address previous patterns without forgetting. In order to learn a new distribution, EWC builds on the assumption that, for an overparameterized model, it exists an optimal configuration of the parameters which is not too distant from the current one in the parameters space and which is able to adapt to upcoming drifts.\\
The importance of each connection is estimated through an approximation of the diagonal of the Fisher Information Matrix, whose computation requires only first-order derivatives. The penalization is implemented by adding a quadratic regularization term to the standard loss function, weighted by the previously computed connection importance. Given task $A$ already learned by the model, the loss function when learning a new task $B$ is:
\begin{equation}
    L(\theta) = L_B(\theta) + \sum_i \frac{\lambda}{2} F_i(\theta_i - \theta_{A,i}^*)^2
\end{equation}
where $F_i$ is the $i$-th diagonal element of the (approximated) Fisher Information Matrix, $L_B$ is the loss for task $B$, $\theta$ are the model parameters, $\theta^*_A$ are the learned parameters for task $A$, $\lambda$ is the hyperparameter controlling the tradeoff between accuracy on old and new task.\\
We implemented EWC in RNNs and we observed its performances in terms of forgetting and accuracy on our benchmarks. At the best of our knowledge, we are the first to study EWC in a sequential context with RNNs.

\subsection{Gated Incremental Memory (GIM)} \label{sec:approach-rnn}
\begin{figure*}
 \centering
 \includegraphics[width=0.9\textwidth]{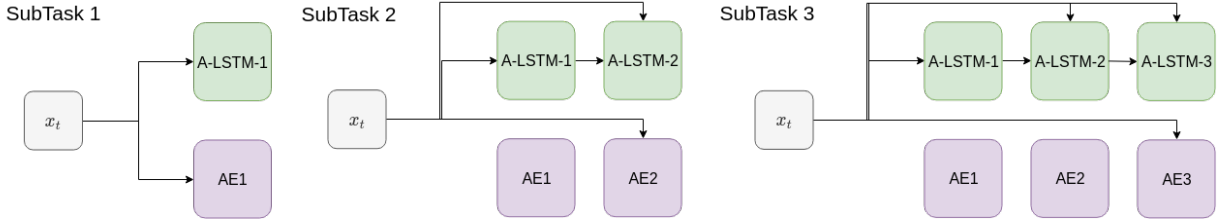}
 \caption{Incremental expansion of the GIM-LSTM during training on $3$ subtasks. When a new subtask is encountered a new module is added.}
 \label{fig:GIM-train}
\end{figure*}
\begin{figure}
 \centering
 \includegraphics[width=0.4\textwidth]{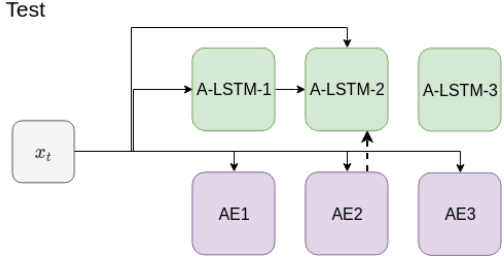}
 \caption{GIM-LSTM at inference time. The input $\mathbf{x}$ is encoded by all the autoencoders. The autoencoder with the minimum reconstruction error (AE2 in the example) determines which module to choose (LSTM-2 in the example). The input is passed to the chosen module to compute the output (dashed line).}
 \label{fig:GIM-test}
\end{figure}

GIM is a general class of dynamic, recurrent architectures that can be built on top of any recurrent model. GIM relies on a progressive memory \cite{rusu_progressive_2016} extension of the underlying RNN model, which uses separate modules for each subtask. It also leverages a set of gating autoencoders, one for each subtask, to automatically select the module that best matches the current input. 
Figures \ref{fig:GIM-train} and \ref{fig:GIM-test} provide an overview of the entire GIM architecture during training and test. 

\subsubsection{RNN Modules}
The main component of GIM is the RNN \textit{module}. As soon as a new distribution arrives, a new RNN module is added on top of the existing architecture and connected to the previous one (Fig. \ref{fig:GIM-train}). The exact inter-modules connections are slightly different depending on the underlying recurrent model. When using the GIM-LSTM, at each timestep, the new module takes as additional input the current hidden state of the previous module. Instead, the GIM-LMN takes as additional input the concatenation of the previous module's memory $h_t^m$ and functional activation $h_t$. These additional inputs allow to easily transfer knowledge from the previous modules to the new ones. To prevent forgetting, when a new module is added to the existing architecture, the previous module's parameters are frozen and no longer updated. Therefore, each module becomes an expert of its own domain. At each timestep $t$, the input vector $x_t$ is forwarded to all modules. Each module has its own output layer and, during training, the last module added to the network is used to generate the final output $y$ from the last hidden state of the module. Given an input $\mathbf{x}=x_1, \hdots, x_T$, the output $y$ for a GIM-LMN with $N$ modules can be computed as follows:
\begin{align*}
 h_{:,1}^m, h_{:,1} &= \text{LMN}_1(\mathbf{x},h_{0,1}^m) \\
 h_{:,j}^m, h_{:,j} &= \text{LMN}_j([\mathbf{x}; h_{:,j-1}^m; h_{:,j-1}] ,h_{0,j}^m), &j=2,...,N \\
 y &= \sigma (W^{mo}_j h^m_{T,N})
\end{align*}
where $\text{LMN}_j$ is the RNN module corresponding to the j-th subtask, $h_{:,j}^m$ and $h_{:,j}$ are the sequences of memory states and functional activations of module $j$ ($:$ indexes all the steps in the sequence), and $[\boldsymbol{\cdot} ; \boldsymbol{\cdot}]$ is the concatenation operator between vectors. The aggregated output $y$ is computed by passing the final memory state $h_{T,N}^m$ through a linear layer. \\
LSTM modules follow the same logic for the forward pass, substituting the final hidden state $h_T$ to the memory state and functional activations of GIM-LMN. A detailed description is provided in Algorithm \ref{alg:forward-alstm} for LSTM modules and Algorithm \ref{alg:forward-almn} for LMN modules.\\

\subsubsection{Gating Autoencoders}
At inference time, GIM models must choose which module to use to compute the output. To solve this problem, each module is associated with an LSTM autoencoder (AE)\cite{srivastava_unsupervised_2015}, which is a sequence-to-sequence model trained to encode and reconstruct the input sequence $\mathbf{x}$. Each autoencoder is trained only on data from the subtask used for the corresponding module. Algorithm \ref{alg:lstm-autoencoder} shows the procedure to reconstruct the input using the AE encoder and decoder.

\subsubsection{Training}
GIM is trained sequentially on each subtask, by adding and training a new module whenever a new task is encountered. The current RNN module is trained by minimizing the Cross Entropy loss for classification tasks, while the corresponding autoencoder is trained to minimize the reconstruction error, by optimizing the mean squared error (MSE) between the input and the reconstructed sequence. The previous modules and autoencoders are not trained anymore and their parameters remain constant. Algorithm \ref{alg:lstm-autoencoder} shows the pseudocode for the training procedure.

\subsubsection{Inference}
At inference time, the computation proceeds in three steps: 
\begin{enumerate}
    \item the autoencoders reconstruct the input sequence;
    \item the subtask is identified by selecting the autoencoder with the minimum reconstruction error;
    \item the module corresponding to the identified subtask is used to compute the output.
\end{enumerate}

The inference procedure is detailed by the following equations:
\begin{align*}
    &\Tilde{\mathbf{x}}_i = \text{AE}_i(\mathbf{x}), i=1,...,N\\
    &k = \argmin_{i} \text{MSE}(\Tilde{\mathbf{x}}_i, \mathbf{x}) \\
    &y = \text{LMN}_k(\mathbf{x}) .
\end{align*}
The algorithms describing the output computation at inference time are in Algorithm \ref{alg:lstm-autoencoder}.\\

\begin{algorithm}[htbp]
    \centering
    \caption{GIM-LSTM Forward Pass for Module $N$}\label{alg:forward-alstm}
    \begin{algorithmic}[1]
        \Function{LSTM-Module-FW}{GIM, $\mathbf{x}$, N}
            \Require GIM-LSTM with at least $N$ modules, $N \geq 1$, $\mathbf{x}$ with $T$ timesteps
            \State $h_{0, 1} \gets 0$  
            \State $h_{:, 1} \gets \text{GIM.LSTM}_1(\mathbf{x}, h_{0, 1})$  
            \For{$d\gets 2, N$}
                \State $h_{0, d} \gets 0$  
                \State $\hat{\mathbf{x}} \gets [ \mathbf{x} ; h_{:, d-1} ]$  
                \State $h_{:, d} \gets \text{GIM.LSTM}_d(\hat{\mathbf{x}}, h_{0, d})$
            \EndFor
            \State $y \gets W^{out} h_{T, N}$
            \State \Return $y$
        \EndFunction
    \end{algorithmic}
\end{algorithm}
\begin{algorithm}[htbp]
    \centering
    \caption{GIM-LMN Forward Pass for Module $N$}\label{alg:forward-almn}
    \begin{algorithmic}[1]
        \Function{LMN-Module-FW}{GIM, $\mathbf{x}$, N}
            \Require GIM-LMN with at least $N$ modules, $N \geq 1$, $\mathbf{x}$ with $T$ timesteps
            \State $h^m_{0, 1} \gets 0$  
            \State $h^m_{:, 1}, h_{:, 1} \gets \text{GIM.LMN}_1(\mathbf{x}, h^m_{0, 1})$  
            \For{$d\gets 2, N$}
                \State $h^m_{0, d} \gets 0$ 
                \State $\hat{\mathbf{x}} \gets [ \mathbf{x} ; h^m_{:, d-1}; h_{:, d-1} ]$
                \State $h^m_{:, d}, h_{:, d} \gets \text{GIM.LMN}_d(\hat{\mathbf{x}}, h^m_{0, d})$
            \EndFor
            \State $y^m_N \gets W^{mo}_N h^m_{T, N}$
            \State \Return $y^m_N$
        \EndFunction
    \end{algorithmic}
\end{algorithm}

\begin{algorithm}
    \caption{Functions to compute the reconstruction of the autoencoder, for training it, and for choosing the GIM module.}
    \begin{algorithmic}[1]
        \Function{Reconstruction}{$\text{AE}$, $\mathbf{x}$}
            \State $h_{enc, 0} \gets 0$
            \State $h_{enc, :} \gets \text{AE.LSTM}_{enc}(\mathbf{x}, h_{enc, 0})$
            \State $h_{dec, :} \gets \text{AE.LSTM}_{dec}(\mathbf{0}, h_{enc, T})$
            \State $\Tilde{\mathbf{x}} \gets \text{AE.}W^{out} h_{dec, :}$
            \State \Return $\Tilde{\mathbf{x}}$
        \EndFunction
        
        \Function{AE-Train}{$\mathcal{D}$}
        \Require $\mid\mathcal{D}\mid > 1$
            \State $l_{ae} \leftarrow []$
            \While{a new distribution $D_k$ is available}
                \State $\text{AE} \gets \text{init-autoencoder}()$
                \State $l_{ae}\text{.append}(\text{AE})$
                \For{training batch $\mathbf{x} \in D_k$}
                    \State $\Tilde{\mathbf{x}} \gets \textsc{Reconstruction}(\text{AE}, \mathbf{x})$
                    \State $J \gets \text{MSE}(\mathbf{x},\Tilde{\mathbf{x}})$
                    \State $\frac{\partial J}{\partial w} \gets \text{backprop}(J)$
                    \State Take a descent step along $\frac{\partial J}{\partial w}$
                \EndFor
            \EndWhile
            \State \Return $l_{ae}$
        \EndFunction
        
        \Function{GIM-Inference}{\text{GIM}, $\mathbf{x}$}
            \State $l_{rec} \gets []$
            \For{ $\text{AE} \in \text{GIM}.l_{ae}$}
                \State $\Tilde{\mathbf{x}} \gets \textsc{Reconstruction}(\text{AE}, \mathbf{x})$
                \State $l_{rec}.append(\text{MSE}(\mathbf{x}, \Tilde{\mathbf{x}}))$
            \EndFor
            \State $m \gets \argmin{l_{rec}}$ \Comment{index of the best autoencoder}
            \State $y \gets \textsc{LSTM-Module-FW}(\text{GIM}, \mathbf{x}, m)$
            \State \Return $y$
        \EndFunction
    \end{algorithmic}
    \label{alg:lstm-autoencoder}
\end{algorithm}

\subsubsection{Advantages of the GIM architecture}
GIM, like Progressive networks, is capable of learning multiple distributions without being affected by forgetting. Freezing old parameters easily guarantees that the model will retain the knowledge about previous subtasks, while the use of the activations of the previous module as additional inputs allow to transfer knowledge from the previous model to the new ones. Additionally, GIM overcomes one of the major drawbacks of Progressive networks \cite{rusu_progressive_2016}: it does not require explicit knowledge about input distributions at test time, since gating autoencoders are able to autonomously recognize the current input and use the appropriate module to compute the output. Compared to Progressive networks, GIM simplifies the inter-modules connections: while Progressive networks use feedforward networks, created between a module and \textit{all} the next ones (quadratic number of connections with respect to the number of modules), GIM employs only concatenation between vectors and connects only adjacent modules. Therefore, the number of connections scales linearly with the number of modules.

\section{Datasets}
We experimented with three different datasets: following the current trend in CL, two of them, MNIST and Devanagari, originate from images. The third one, Audioset, is constructed by processing short clips of audio sounds and is therefore more representative of a sequential, dynamic environment\footnote{Code to reproduce results and Audioset classes used in the experiments can be found at \url{https://github.com/AndreaCossu/ContinualLearning-SequentialProcessing}}. \\
MNIST and Devanagari are adapted to sequential data processing by transforming each image in a sequence of pixels, which are then shuffled according to a fixed, random permutation. Permuting the images ensures that the RNN performance is not affected by the long sequence of non-informative elements (pixels) which are present at the end of each sequence. Considering the image resolution of $28$x$28$, the sequences consist of $784$ timesteps, making these datasets challenging not only for the CL scenario but also for recurrent models in general due to the length of the input sequences. \\
In order to create dynamic environments, we choose to follow the standard approach of CL literature \cite{sodhani_training_2018, schak_study_2019} by dividing each dataset into groups of non-overlapping classes, which we call subtasks. When training on a specific dataset (i.e. when addressing a particular task), subtasks are presented to the model sequentially one after the other (the next one starts when the previous has ended). \\
Concept drift is present on the output layer of each model since the input distribution associated with each output unit changes from one subtask to the next. The model should adapt to the drift and learn the new concept without forgetting the previous ones. When finished training on all subtasks, the model is tested on both last and previous subtasks to assess its resilience to forgetting.\\

\subsection{MNIST}
One of the most used datasets in Machine Learning and CL is the MNIST dataset of handwritten digit \cite{lecun_gradient-based_1998}. Each image has size $28$x$28$, gray scaled, which leads to input sequences with $28 \cdot 28 = 784$ scalars. We created $5$ subtasks, corresponding to the $5$ digits partitions: $(0,1), (2,3), (4,5), (6,7), (8,9)$ and we trained the models to classify each pair of digits, switching to the next subtasks once the previous one is completed. \\

\subsection{Devanagari}
Devanagari is a dataset composed of images of handwritten characters belonging to $46$ different classes \cite{acharya_deep_2015}. Each class has $1,700$ images, each of which of size $32$x$32$, gray-scaled. Following the approach of \cite{schak_study_2019}, we randomly selected $10$ classes out of the $46$. In addition, we also remove the padding along the borders of the image, resulting in $28$x$28$ gray-scaled input images. The total length of the input sequence is therefore the same as in MNIST.\\
We split the selected classes in $5$ subtasks of $2$ classes each: (\textit{gha}, \textit{cha}), (\textit{chha}, \textit{daa}), (\textit{bha}, \textit{ma}), (\textit{motosaw}, \textit{petchiryakha}), (\textit{1}, \textit{3}). The last subtask is composed of digits, which however have a completely different representation than the ones in MNIST. The objective of the task is to assign to each sequence the correct class. 
\subsection{Audioset}
Audioset \cite{gemmeke_audio_2017} is a collection of annotated audio events, extracted from $10$ seconds audio clips and organized hierarchically in classes. The objective is the classification of a sound from its audio clip source, embedded through a VGG-acoustic model into $10$ vectors, one per second, each of dimension $128$. To implement a CL scenario, we selected $40$ audio classes and split them among $4$ subtasks ($10$ classes per subtask). We selected the $40$ classes according to the procedure outlined by \cite{kemker_measuring_2018}. Since the authors did not publish the classes, we randomly selected them from the superset resulting from their preprocessing pipeline.
Audioset data has already been used in literature to assess CL skills \cite{kemker_measuring_2018}. However, the authors focused on the task from a \emph{static} perspective, relying on the use of feedforward models only. Since the preprocessing step provides, for each audio clip, a sequence of fixed-size embeddings, it is possible to concatenate the vectors into a single large vector and feed it to the network. The sequential aspect of the task, however, is completely lost. At the best of our knowledge, we are the first to tackle Audioset in CL scenarios with recurrent models. It is also important to notice that the task difficulty is increased when using recurrent networks, since the model is not able to see the input in its entirety (like in feedforward networks), but it has to scan it one timestep at a time.\\

\begin{figure*}[htbp]
\centering
\begin{subfigure}{0.24\textwidth}
\includegraphics[width=0.95\linewidth]{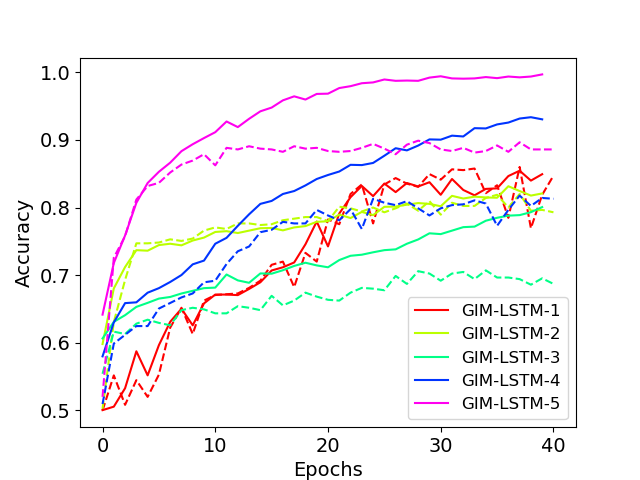}
\caption{GIM-LSTM on Devanagari: $40$ epochs}
\end{subfigure}
\begin{subfigure}{0.24\textwidth}
\includegraphics[width=0.95\linewidth]{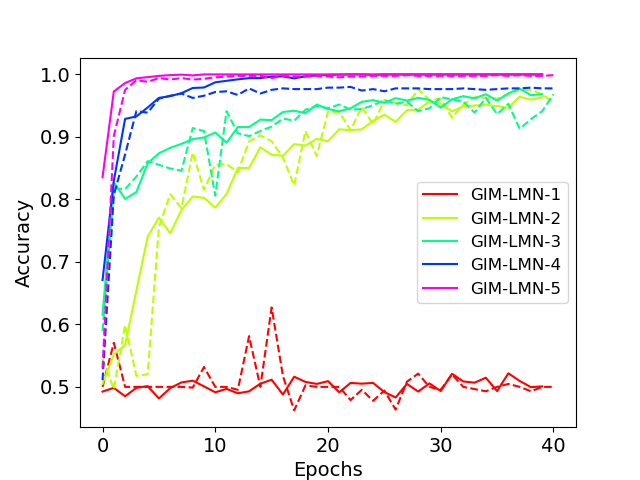}
\caption{GIM-LMN on Devanagari: $40$ epochs}
\end{subfigure}
\begin{subfigure}{0.24\textwidth}
\includegraphics[width=0.95\linewidth]{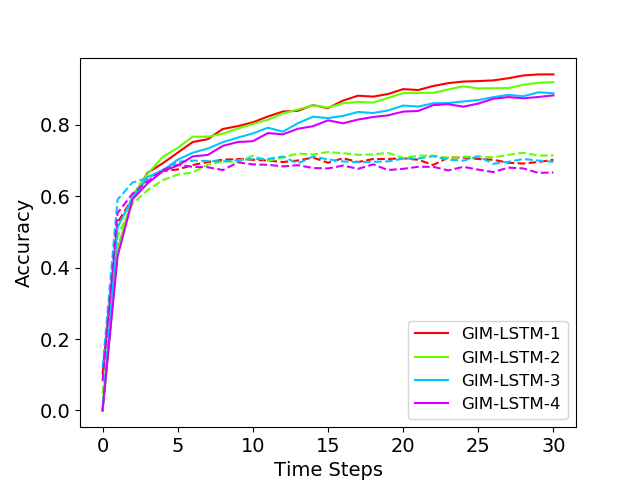}
\caption{GIM-LSTM on Audioset: $30,000$ time steps, printed every $1000$ steps.}
\end{subfigure}
\begin{subfigure}{0.24\textwidth}
\includegraphics[width=0.95\linewidth]{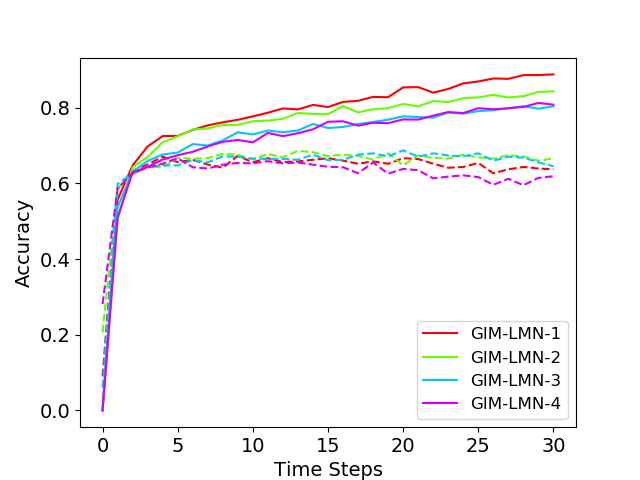}
\caption{GIM-LMN on Audioset: $30,000$ time steps, printed every $1000$ steps.}
\end{subfigure}
\label{fig:lc}
\caption{Examples of accuracy curves on training set (solid line) and validation set (dashed line).}
\end{figure*}

\section{Experiments}
On MNIST and Devanagari we trained all models with Adam optimizer \cite{kingma_adam:_2015}, learning rate of $1e-4$, mini batch size of $32$. The number of hidden units is set to $128$ for both LSTM-based and LMN-based (functional and memory component) models. \\
It is well known that orthogonal initialization of memory weight matrixes can improve learning when dealing with long sequences and linear memories \cite{mhammedi_efficient_2017, henaff_recurrent_2016, vorontsov_orthogonality_2017}. Therefore, we chose to use such initialization for LMN-based models and also to preserve it during training through an additional penalty in the loss function, expressed by:
\begin{equation}
    \beta \lVert (W^{mm})^T W^{mm} - I \rVert^2
\end{equation}
where $W^{mm}$ is the memory weight matrix of the LMN, $I$ is the identity matrix and $\beta$ is the hyperparameter associated to the penalization.\\
We used a value of $\beta = 0.1$, which was capable of preserving orthogonality on all experiments.\\
On Audioset we adopted a different configuration by using smaller models with $16$ hidden units on LSTM and LMN (memory and functional component). This choice was determined by the limited amount of data points available for each class in Audioset: larger models led to overfitting without any performance improvements. We used the RMSProp optimizer, with learning rate of $3e-5$, momentum of $0.9$ and L2 regularization with hyperparameter of $1e-3$ and mini batch size of $4$.\\
LSTM autoencoders use $500$ hidden units on encoder and decoder ($36.22\%$ compression on MNIST and Devanagari, $60.94\%$ on Audioset) trained with Adam optimizer and learning rate of $1e-4$. On Audioset, we also use L2 regularization with hyperparameter $1e-3$. \\

We compare GIM-LSTM and GIM-LMN with standard LSTM and LMN and with the popular EWC method \cite{kirkpatrick_overcoming_2017}. EWC requires to choose the value of the hyperparameter regulating the tradeoff between new incoming subtask and older ones. We choose the value of $0.4$ out of $0.01, 0.1, 0.4, 1.0$, since it gave the best performances on a held-out validation set. \\

Table \ref{tab:results} provides the results of our experiments. Paired plots (Fig. 4) show the comparison between the validation performance for each subtask computed after training on that subtask (on the left) and the test performance computed after the final training is completed for all the subtasks. Therefore, the forgetting for each configuration can be evaluated by looking at the difference between the left and right points for each subtask.  \\
MNIST and Devanagari are binary classification tasks, hence a random classifier would score $0.50$ accuracy on all subtasks. On Audioset, being composed of $10$ classes per subtask, a random classifier would score $0.10$ accuracy on all subtasks. \\
We also show examples of learning curves, comparing GIM-LMN with GIM-LSTM (Fig. 3). Audioset shows early overfitting even with small models. A similar behavior, even if less drastic, is detected on Devanagari. It is, however, important to stress the fact that the learning curves cannot show the effect of forgetting because they are computed using the data from the current subtask, while we are interested in the final test accuracy, measured after training on all subtasks. \\

\begin{table*}[htbp]
\caption{Validation/Test accuracy on all datasets (D) and subtasks (S). Validation accuracy for a subtask computed right after training on that specific subtask. Test accuracy computed at the end of training on all subtasks.}
\label{tab:results}
\begin{center}
\begin{tabular}{| c | c | c | c | c | c | c | c |}
\hline
\textbf{D} & \textbf{S} & \textbf{LSTM} & \textbf{LMN} & \textbf{EWC-LSTM} & \textbf{EWC-LMN} & \textbf{GIM-LSTM} & \textbf{GIM-LMN} \\
\hline
\hline

\multicolumn{1}{|c|}{\multirow{5}{*}{\rotatebox[origin=c]{90}{MNIST}}}      & \multicolumn{1}{c|}{1} & \multicolumn{1}{c|}{ $0.97$ / $0.55$} & \multicolumn{1}{c|}{$0.99$ / $0.45$} & \multicolumn{1}{c|}{$0.95$ / $0.74$} & \multicolumn{1}{c|}{$0.98$ / $0.32$} & \multicolumn{1}{c|}{ $0.97$ / $0.97$} & \multicolumn{1}{c|}{$0.99$ / $\mathbf{0.98}$} \\ \cline{2-8} 
\multicolumn{1}{|c|}{}                            & \multicolumn{1}{c|}{2} & \multicolumn{1}{c|}{$0.86$ / $0.47$} & \multicolumn{1}{c|}{$0.97$ / $0.58$} & \multicolumn{1}{c|}{$0.72$ / $0.54$} & \multicolumn{1}{c|}{$0.88$ / $0.63$} & \multicolumn{1}{c|}{$0.92$ / $0.50$} & \multicolumn{1}{c|}{$0.98$ / $\mathbf{0.98}$} \\ \cline{2-8} 
\multicolumn{1}{|c|}{}                            & \multicolumn{1}{c|}{3} & \multicolumn{1}{c|}{$0.94$ / $0.18$} & \multicolumn{1}{c|}{ $0.99$ / $0.14$} & \multicolumn{1}{c|}{$0.62$ / $0.43$} & \multicolumn{1}{c|}{ $0.80$ / $\mathbf{0.45}$} & \multicolumn{1}{c|}{$0.93$ / $0.35$} & \multicolumn{1}{c|}{$0.99$ / $0.35$} \\ \cline{2-8} 
\multicolumn{1}{|c|}{}                            & \multicolumn{1}{c|}{4} & \multicolumn{1}{c|}{$0.98$ / $0.75$} & \multicolumn{1}{c|}{$0.99$ / $0.76$} & \multicolumn{1}{c|}{$0.56$ / $0.54$} & \multicolumn{1}{c|}{$0.93$ / $0.76$} & \multicolumn{1}{c|}{$0.96$ / $0.91$} & \multicolumn{1}{c|}{$0.99$ / $\mathbf{0.96}$} \\ \cline{2-8} 
\multicolumn{1}{|c|}{}                            & \multicolumn{1}{c|}{5} & \multicolumn{1}{c|}{$0.94$ / $0.94$} & \multicolumn{1}{c|}{$0.98$ / $0.98$} & \multicolumn{1}{c|}{$0.65$ / $0.64$} & \multicolumn{1}{c|}{$0.71$ / $0.72$} & \multicolumn{1}{c|}{$0.88$ / $0.76$} & \multicolumn{1}{c|}{$0.97$ / $\mathbf{0.95}$} \\
\cline{2-8} 
\multicolumn{1}{|c|}{}                            & \multicolumn{1}{c|}{\textbf{Mean}} & \multicolumn{1}{c|}{$0.94$ / $0.58$} & \multicolumn{1}{c|}{$0.98$ / $0.58$} & \multicolumn{1}{c|}{$0.70$ / $0.58$} & \multicolumn{1}{c|}{$0.86$ / $0.58$} & \multicolumn{1}{c|}{$0.93$ / $0.70$} & \multicolumn{1}{c|}{$0.98$ / $\mathbf{0.84}$} \\ \hline \hline

\multicolumn{1}{|c|}{\multirow{5}{*}{\rotatebox[origin=c]{90}{Devanagari}}}  & \multicolumn{1}{c|}{1} & \multicolumn{1}{c|}{ $0.82$ / $0.48$} & \multicolumn{1}{c|}{$0.50$ / $0.55$} & \multicolumn{1}{c|}{$0.80$ / $\mathbf{0.60}$} & \multicolumn{1}{c|}{$0.50$ / $0.52$} & \multicolumn{1}{c|}{ $0.82$ / $0.59$} & \multicolumn{1}{c|}{$0.50$ / $0.39$} \\ \cline{2-8} 
\multicolumn{1}{|c|}{}                            & \multicolumn{1}{c|}{2} & \multicolumn{1}{c|}{$0.76$ / $0.44$} & \multicolumn{1}{c|}{$0.85$ / $0.59$} & \multicolumn{1}{c|}{$0.76$ / $0.48$} & \multicolumn{1}{c|}{$0.50$ / $0.49$} & \multicolumn{1}{c|}{$0.81$ / $\mathbf{0.74}$} & \multicolumn{1}{c|}{$0.96$ / $0.73$} \\ \cline{2-8} 
\multicolumn{1}{|c|}{} & \multicolumn{1}{c|}{3} & \multicolumn{1}{c|}{$0.65$ / $0.49$} & \multicolumn{1}{c|}{ $0.87$ / $0.69$} & \multicolumn{1}{c|}{$0.63$ / $0.55$} & \multicolumn{1}{c|}{ $0.49$ / $0.56$} & \multicolumn{1}{c|}{$0.71$ / $0.67$} & \multicolumn{1}{c|}{$0.95$ / $\mathbf{0.86}$} \\ \cline{2-8} 
\multicolumn{1}{|c|}{}                            & \multicolumn{1}{c|}{4} & \multicolumn{1}{c|}{$0.76$ / $0.49$} & \multicolumn{1}{c|}{$0.98$ / $0.35$} & \multicolumn{1}{c|}{$0.71$ / $\mathbf{0.60}$} & \multicolumn{1}{c|}{$0.50$ / $0.48$} & \multicolumn{1}{c|}{$0.79$ / $0.51$} & \multicolumn{1}{c|}{$0.98$ / $0.33$} \\ \cline{2-8} 
\multicolumn{1}{|c|}{}                            & \multicolumn{1}{c|}{5} & \multicolumn{1}{c|}{$0.90$ / $0.91$} & \multicolumn{1}{c|}{$0.99$ / $0.99$} & \multicolumn{1}{c|}{$0.86$ / $0.82$} & \multicolumn{1}{c|}{$0.51$ / $0.52$} & \multicolumn{1}{c|}{$0.87$ / $0.66$} & \multicolumn{1}{c|}{$0.99$ / $\mathbf{0.83}$} \\
\cline{2-8} 
\multicolumn{1}{|c|}{}                            & \multicolumn{1}{c|}{\textbf{Mean}} & \multicolumn{1}{c|}{$0.78$ / $0.56$} & \multicolumn{1}{c|}{$0.84$ / $0.63$} & \multicolumn{1}{c|}{$0.75$ / $0.61$} & \multicolumn{1}{c|}{$0.50$ / $0.51$} & \multicolumn{1}{c|}{$0.80$ / $\mathbf{0.63}$} & \multicolumn{1}{c|}{$0.88$ / $\mathbf{0.63}$} \\ \hline \hline

\multicolumn{1}{|c|}{\multirow{4}{*}{\rotatebox[origin=c]{90}{Audioset}}}   & \multicolumn{1}{c|}{1} & \multicolumn{1}{c|}{$0.63$ / $0.10$ } & \multicolumn{1}{c|}{ $0.60$ / $0.05$} & \multicolumn{1}{c|}{$0.67$ / $0.08$} & \multicolumn{1}{c|}{$0.61$ / $0.04$} & \multicolumn{1}{c|}{$0.68$ / $\mathbf{0.64}$} & \multicolumn{1}{c|}{$0.62$ / $0.55$} \\ \cline{2-8} 
\multicolumn{1}{|c|}{}                            & \multicolumn{1}{c|}{2} & \multicolumn{1}{c|}{$0.71$ / $0.09$} & \multicolumn{1}{c|}{ $0.67$ / $0.08$} & \multicolumn{1}{c|}{$0.71$ / $0.12$} & \multicolumn{1}{c|}{$0.65$ / $0.14$} & \multicolumn{1}{c|}{$0.73$ / $\mathbf{0.71}$} & \multicolumn{1}{c|}{$0.68$ / $0.65$} \\ \cline{2-8} 
\multicolumn{1}{|c|}{}                            & \multicolumn{1}{c|}{3} & \multicolumn{1}{c|}{$0.68$ / $0.13$} & \multicolumn{1}{c|}{ $0.64$ / $0.14$} & \multicolumn{1}{c|}{$0.68$ / $0.13$} & \multicolumn{1}{c|}{ $0.64$ / $0.15$} & \multicolumn{1}{c|}{$0.71$ / $\mathbf{0.57}$} & \multicolumn{1}{c|}{$0.63$ / $0.54$} \\ \cline{2-8} 
\multicolumn{1}{|c|}{}                            & \multicolumn{1}{c|}{4} & \multicolumn{1}{c|}{$0.67$ / $0.46$} & \multicolumn{1}{c|}{$0.63$ / $0.43$} & \multicolumn{1}{c|}{$0.62$ / $0.50$} & \multicolumn{1}{c|}{$0.62$ / $0.47$} & \multicolumn{1}{c|}{$0.67$ / $\mathbf{0.50}$} & \multicolumn{1}{c|}{$0.63$ / $0.42$} \\
\cline{2-8} 
\multicolumn{1}{|c|}{}                            & \multicolumn{1}{c|}{\textbf{Mean}} & \multicolumn{1}{c|}{$0.67$ / $0.20$} & \multicolumn{1}{c|}{$0.64$ / $0.18$} & \multicolumn{1}{c|}{$0.67$ / $0.21$} & \multicolumn{1}{c|}{$0.63$ / $0.20$} & \multicolumn{1}{c|}{$0.70$ / $\mathbf{0.61}$} & \multicolumn{1}{c|}{$0.64$ / $0.54$} \\ \hline

\end{tabular}
\end{center}
\end{table*}

\begin{figure*}[htbp]
\centering
\begin{subfigure}{0.32\textwidth}
\includegraphics[width=0.99\linewidth]{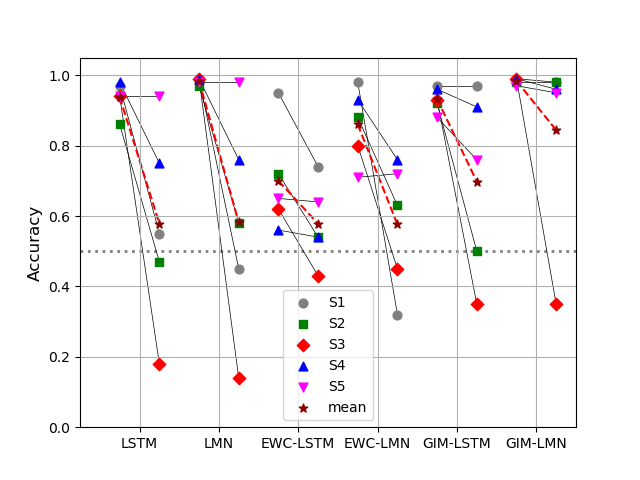}
\caption{MNIST}
\end{subfigure}
\begin{subfigure}{0.32\textwidth}
\includegraphics[width=0.99\linewidth]{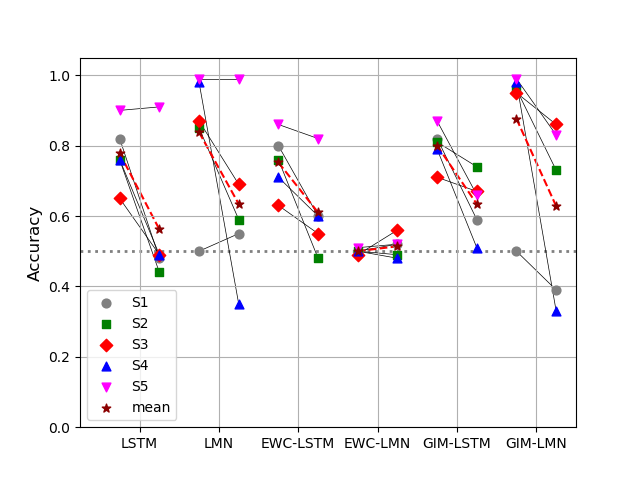}
\caption{Devanagari}
\end{subfigure}
\begin{subfigure}{0.32\textwidth}
\includegraphics[width=0.99\linewidth]{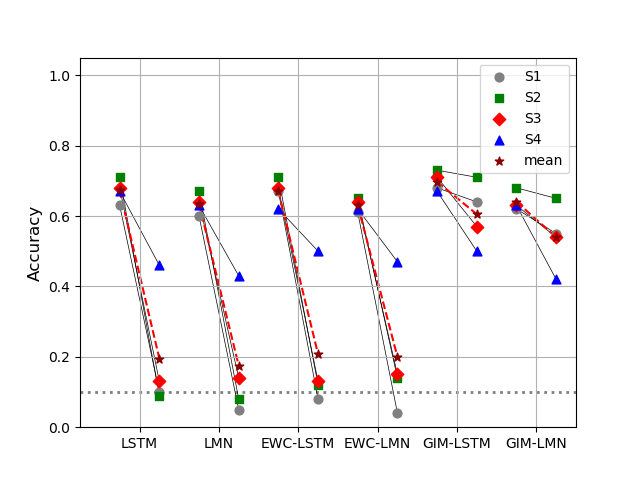}
\caption{Audioset}
\end{subfigure}
\label{fig:paired}
\caption{Paired plots for the three tasks. Each pair plot shows, for each model, validation accuracy (left point) computed for each subtask after training on that subtask, and test accuracy (right point) computed for each subtask at the end of the entire training. For each subtask, validation and test accuracies are connected by a line. Therefore, the drop in performance due to forgetting is the difference between the two points. The red dashed line is the average among subtasks. Horizontal dotted line is equal to random classifier performance ($0.5$ for MNIST and Devanagari, $0.1$ for Audioset).}
\end{figure*}

\section{Discussion}
Table \ref{tab:results} and Figure 4 show that, in accordance with the results presented in \cite{schak_study_2019}, LSTM and LMN models suffer from catastrophic forgetting of old subtasks, independently of the performance achieved on the validation set during training. Even models regularized with EWC are not able to mitigate catastrophic forgetting. Notably, EWC always lowers the performance of the last subtask, an effect that is probably caused by the strong regularization imposed on the model weights. We were unable to find an EWC setting capable of guaranteeing a good tradeoff between current and previous subtasks accuracies. We hypothesize that the recurrence in RNNs could be the cause of the poor performance of EWC, leading to an importance evaluation through the Fisher Information matrix which is not representative of the (sub)task on which the model is trained. However, further studies will be needed to validate or contradict this hypothesis.\\
GIM models are by far the best performing ones, since they successfully learn on dynamic environments while limiting forgetting. \\
On Audioset, GIM-LSTM and GIM-LMN are capable of maintaining comparable performance on all subtasks once training is finished, while performance for standard and EWC-based models drops below the random baseline for some subtasks. This means that the autoencoders successfully recognize the incoming distribution and select the correct module to produce the output without any information on the incoming input labels. \\
On MNIST we obtained similar results, with some exceptions: GIM-LSTM underwent complete forgetting on $2$ subtasks out of $5$, while GIM-LMN experienced it on $1$ out of $5$. In those cases, autoencoders failed to reconstruct the input sequences, leading to the choice of the wrong module for the final classification. On these subtasks, the EWC version of LMN and LSTM surpassed GIM architectures.\\
Devanagari is the most challenging task: GIM models still exhibited complete forgetting on $2$ subtasks out of $5$, with reduced performances on almost all the others. However, this behavior is common to all models on this task. EWC-LMN does not show significant differences between performances on the validation set and final test only because it is unable to learn the task, achieving an accuracy equivalent to the one of a random classifier. \\

\section{Conclusions}
The main objective of this work is to draw attention to the problem of CL for sequential data processing by introducing GIM, a recurrent CL architecture inspired by the Progressive networks \cite{rusu_progressive_2016}. Our benchmarks show that GIM is able to mitigate forgetting on computer vision and sequential audio data. GIM surpasses LSTM, LMN and their corresponding EWC version on the large majority of the experiments. The performance of GIM models depends on the reconstruction error of the subtask's autoencoders. In the future, different models for the autoencoders could be used to further improve the performance. The comparison with EWC on sequential, dynamic environments supports the claim that recurrent architectures need to be adapted to manage CL scenarios and encourages future works towards an in-depth study of their behaviors.\\

\bibliography{main}
\bibliographystyle{ieeetr}

\end{document}